\documentclass[conference,compsoc]{IEEEtran}
\usepackage[nocompress]{cite}
\usepackage[pdftex]{graphicx}
\usepackage{amsmath}
\usepackage{algorithmic}

\usepackage{algorithm}
\usepackage{array}
\usepackage[caption=false,font=footnotesize,labelfont=sf,textfont=sf]{subfig}
\usepackage[caption=false,font=footnotesize,labelfont=sf,textfont=sf]{subfig}

\usepackage{fixltx2e}
\usepackage{url}

\usepackage{amsthm}
\usepackage{amsfonts}
\usepackage{amssymb}
\usepackage{booktabs}
\usepackage{xcolor}
\theoremstyle{plain}
\newtheorem{theorem}{Theorem}[section]
\newtheorem{proposition}[theorem]{Proposition}

\newtheorem{corollary}[theorem]{Corollary}
\theoremstyle{definition}
\newtheorem{definition}[theorem]{Definition}

\theoremstyle{remark}

\DeclareMathOperator{\R}{\mathbb{R}}

\DeclareMathOperator{\E}{\mathbb{E}}
\DeclareMathOperator{\Model}{\mathcal{M}}

\DeclareMathAlphabet\mathbfcal{OMS}{cmsy}{b}{n}

\newcommand{\Prob}{\mathbb{P}}
\newcommand{\Probof}[1]{\mathbb{P}\left[#1\right]}

\newcommand{\obfs}[1]{\tilde{#1}}
\newcommand{\obfsX}{\obfs{X}}
\newcommand{\obfsx}{\obfs{x}}

\newcommand{\Expsub}[2]{\E_{#1}\left[#2\right]}

\DeclareMathOperator{\Loss}{\mathcal{L}}
\DeclareMathOperator{\LossO}{\mathcal{L}_O}
\DeclareMathOperator{\LossU}{\mathcal{L}_U}
\DeclareMathOperator{\LossMI}{\mathcal{L}_O^{MI}}
\DeclareMathOperator{\LossACS}{\mathcal{L}_O^{ACS}}
\DeclareMathOperator{\LossMNP}{\mathcal{L}_O^{MNP}}
\DeclareMathOperator{\LossCGE}{\mathcal{L}_O^{CGE}}
\DeclareMathOperator{\LossBCS}{\mathcal{L}_O^{BCS}}

\newcommand{\pcat}{+Cat}
\newcommand{\sgtmi}{MI}

\newcommand{\sgtlog}{Comp. Entr.\,}
\newcommand{\sgtlogcat}{\sgtlog\pcat\,}
\newcommand{\sgtcos}{CosSim\,}
\newcommand{\sgtcoslog}{\sgtcos + \sgtlog\,}
\newcommand{\constnoise}{Gaussian\,}
\newcommand{\sgtabscos}{AbsCos\,}
\newcommand{\sgtmiabscos}{\sgtmi + \sgtabscos}
\newcommand{\sgtnormpen}{Norm}
\newcommand{\sgtfull}{\sgtmiabscos + \sgtnormpen}

\DeclareMathOperator{\rank}{\text{rank}}

\theoremstyle{definition}

\begin{document}
\title{Learning Obfuscations Of LLM Embedding Sequences: \\ Stained Glass Transform}

\author{
\IEEEauthorblockN{Jay Roberts}
\IEEEauthorblockA{Protopia AI\\
jay@protopia.ai}
\and
\IEEEauthorblockN{Kyle Mylonakis}
\IEEEauthorblockA{Protopia AI\\
kyle@protopia.ai}
\and
\IEEEauthorblockN{Sidhartha Roy}
\IEEEauthorblockA{Protopia AI\\
sid@protopia.ai}
\and
\IEEEauthorblockN{Kaan Kale}
\IEEEauthorblockA{Protopia AI\\
kaan.kale@protopia.ai}

}


% compsoc conferences compared to traditional conferences), use this
%Telephone: (800) 555--1212, Fax: (888) 555--1212}


\maketitle

%   Abstract
% -------------------------------------

\begin{abstract}
The high cost of ownership of AI compute infrastructure and challenges of robust serving of large language models (LLMs) has led to a surge in managed Model-as-a-service deployments. Even when enterprises choose on-premises deployments, the compute infrastructure is typically shared across many teams in order to maximize the return on investment. In both scenarios the deployed models operate only on plaintext data, and so enterprise data owners must allow their data to appear in plaintext on a shared or multi-tenant compute infrastructure. This results in data owners with private or sensitive data being hesitant or restricted in what data they use with these types of deployments. In this work we introduce the Stained Glass Transform, a learned, stochastic, and sequence dependent transformation of the word embeddings of an LLM which information theoretically provides privacy to the input of the LLM while preserving the utility of model. We theoretically connect a particular class of Stained Glass Transforms to the theory of mutual information of Gaussian Mixture Models. We then calculate a-postiori privacy estimates, based on mutual information, and verify the privacy and utility of instances of transformed embeddings through token level metrics of privacy and standard LLM performance benchmarks.
\end{abstract}


% \fi
\IEEEpeerreviewmaketitle

% -------------------------------------
%   Intro
% -------------------------------------
\section{Introduction}
\begin{figure}[h]
    \centering
    \includegraphics[width=0.95\linewidth]{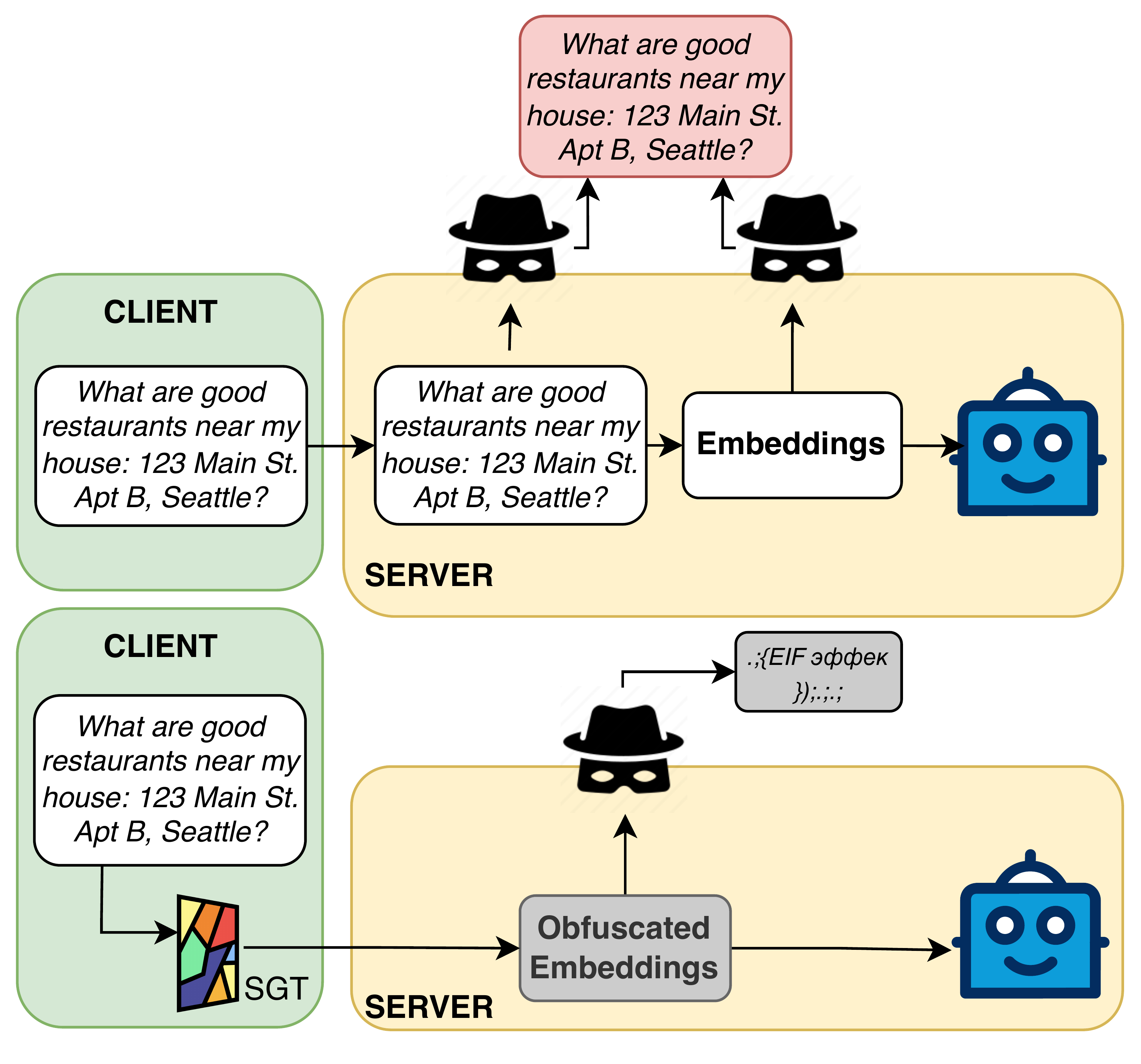}
    \caption{Client input is sent as plaintext (top) and intercepted by an adversary. The points of potential leak include not only the tokens but also the target model's embedding which can be trivially inverted back to text with a nearest neighbor attack. With the Stained Glass Transform (bottom) the Client no longer transmits plaintext and instead the SGT converts the plaintext into a protected embedding that can be directly used by the LLM while still maintaining privacy.}
    \label{fig:high level}
\end{figure}

% Advances in machine learning over the past several years have fundamentally transformed how enterprises and consumers interact with their data. 
The rise of Transformers and Large Language Models (LLMs) has driven rapid advances across a wide range of technologies, including code completion, retrieval-augmented generation (RAG), and other systems enabling natural language interaction with vast amounts of data \cite{ HUSEIN2025103917, 10.5555/3495724.3496517}.
%The rise of Transformers and Large Language Models (LLMs) allowed for the rapid development of technologies such as retrieval augmented generation (RAG) which have allowed for the natural language interaction with vast amounts of data \cite{10.5555/3495724.3496517}. Enterprises attempting to adopt these technologies face data owners who are hesitant to send their sensitive data to untrusted parties. Yet this usage model of sensitive data is often required, as in the case of enterprises using third-party LLM providers in the cloud, or in the case of enterprises using on premises shared resources across stove-piped organizations to lower the costs of the deployed technology and increase the return on investment of the communal compute resources.
% While much attention has been given for the private training of these systems with techniques such as federated learning and differentially private stochastic gradient descent, the inference time privacy of LLMs remains largely open. 
While the privacy of training data of LLMs is well studied, enterprises looking to privately use pretrained LLMs for inference are limited to technologies such as tokenization, masking, and abstraction \cite{mireshghallah2021privacyregularizationjointprivacyutility,siyan2025papillonprivacypreservationinternetbased}. Unfortunately such methods are insufficient as they can significantly degrade LLM performance when the sensitive data is critical to the inference and are hard to use when it is unclear what is sensitive in the prompt \cite{MARTINEZ2012304,10448234,10.1007/s10462-022-10204-6}. In some cases, the solutions used to identify sensitive data use LLM functionalities of their own that require the data owner to trust the tokenization/masking provider with access to their plain-text sensitive information in order to protect it from the original provider (e.g. Amazon Comprehend, Azure AI Language, Google Cloud Sensitive Data Protection). 

% In this paper we ask:

% \textit{Is it possible to information theoretically protect the input embeddings of a fixed LLM at inference time without impacting the LLM's utility?}

The inference attack vector we address is one where the the client's data has left their trust zone to be used by a LLM where it necessarily must be in a unencrypted state. At this point we assume
an unauthorized user is able to access the client's unencrypted data by way of access to the server running the AI model (see Figure \ref{fig:high level}).

We must then determine what data can be sent out of the trust zone that will maintain the client's privacy, but also be effectively used by the LLM. Transmitting the tokens instead of text provides no protection. The next representation that could be sent are the embeddings of the tokens rather than the tokens themselves. These are not immediately human readable and would not change the model performance at all, however, they can be trivially inverted back into plaintext, even in the presence of Gaussian noise, using nearest neighbor searches (see Figure \ref{fig:high level}) \cite{morris2023textembeddingsrevealalmost}. 

In this paper we seek an answer to the following question: \textit{Is it possible to information theoretically protect the input embeddings sent to a fixed LLM at inference time without greatly impacting the LLM's utility?}

%The attack vector we are interested in mitigating is one where the input prompts (after being transformed into embeddings) have left the trust zone of an enterprise. An adversary with access to such embeddings effectively has access to the plaintext prompts (see Figure \ref{fig:high level}). This vulnerability stems from the fact that the decryption of the embeddings of the prompt necessarily occurs on the LLM side, as LLMs operate on plaintext data. This leaves the plaintext prompts susceptible to being stored inadvertently and exposed to anyone who gains access to the server, including the vendor or other tenants in multi-tenant deployments. This highlights the need for a more secure approach to handling input prompts and their embeddings, particularly in scenarios where data is stored on non-private systems after network encryption but prior to processing by the LLM, or where observability and inference technologies for monitoring and managing the inputs to the LLM are deployed. 

Our affirmative answer to this is supported by the introduction of the Stained Glass Transform (SGT) for LLMs, a stochastic, input, and sequence dependent learned transformation of an LLM's token embeddings (Figure \ref{fig:sgt-block-diagram}). 
%Token embeddings of an LLM are in one to one correspondence with the vocabulary of the tokenizer and thus should be considered as sensitive as their corresponding token. 
Working with vector token embeddings avoids the pitfalls of tokenization based privacy mentioned above by allowing us to apply continuous optimization techniques rather than limiting ourselves to discrete or combinatorial optimization. 

We organize the remaining sections as follows: Section \ref{sec:background} provides background, relevant related works, and notation for the rest of the paper. Then Section \ref{sec:sgt} introduces the Stained Glass Transform.
Section \ref{sec:learning-prob} formulates the learning problem used to define the SGT. 
Privacy metrics are introduce in Section \ref{sec:metrics}, and experimental results are reported in Section \ref{sec:experiments}. 
% We offer concluding remarks in Section \ref{sec:conclusion}.

% -------------------------------------
%   Background
% -------------------------------------
\section{Background}\label{sec:background}

% -------------------------------------
%   Large Language Models
% -------------------------------------
\subsection{Large Language Models} Though the Stained Glass Transform (SGT) is modality agnostic (see Definition \ref{def:sgt}), in this work we focus on how the SGT is used to protect the inputs of decoder-only LLMs ( the most common architecture of state-of-the-art autoregressive LLMs \cite{10.5555/3295222.3295349, Radford2018ImprovingLU}). This subsection provides background on how these LLMs function. 

Broadly, current state-of-the-art LLM architectures can be broken into four stages: tokenization, embedding, decoding, and sampling.
The tokenization stage converts text into a pre-defined vocabulary of tokens. These tokens are then mapped through an embedding function, which is a {one-to-one} function mapping token ids to embedding vectors. Often we refer to the embedding function as the embedding table or embedding matrix. Once the sequence of text has been converted to sequences of embeddings, the embeddings are fed through attention based decoder layers to generate a conditional probability distribution of the next token given the input embeddings. A sampling method such as beam search is then used to produce a sample of the predicted next token, which is then appended to the original input tokens and the process continues auto-regressively until a termination criteria is hit \cite{Radford2018ImprovingLU, Radford2019LanguageMA}.

Importantly, unless the tokenizer and embedding tables are secret,
\emph{knowledge of clean token embeddings is equivalent to knowledge of text}. The goal of the SGT for LLMs is to protect these clean token embeddings and corresponding original prompts while maintaining the performance of the LLM. 

\begin{figure*}[!t]
\centering
\subfloat[Obfuscating Inputs With SGT]{
    \includegraphics[width=0.4\textwidth]{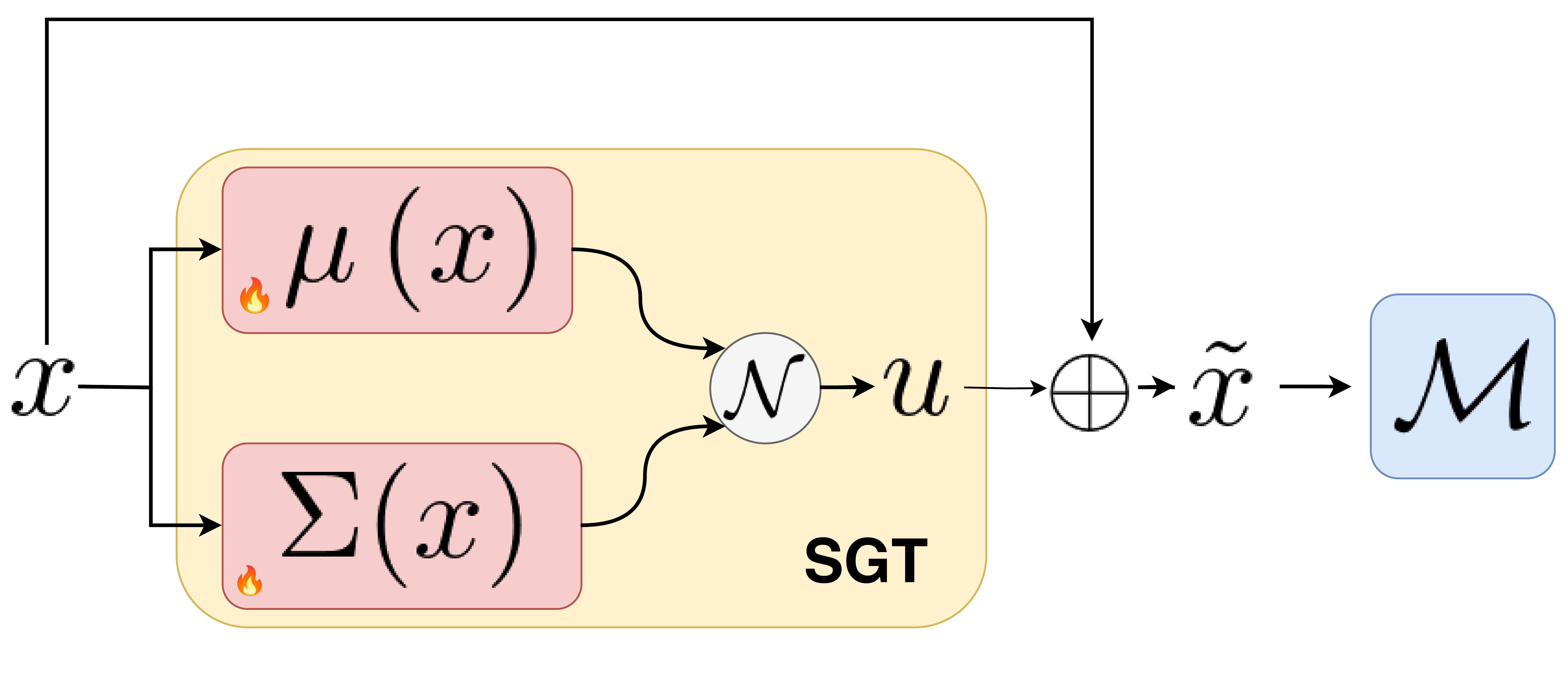}%
    \label{fig:sgt-block-inf}
}
\hfil
\subfloat[Training SGT Model]{
    \includegraphics[width=0.4\textwidth]{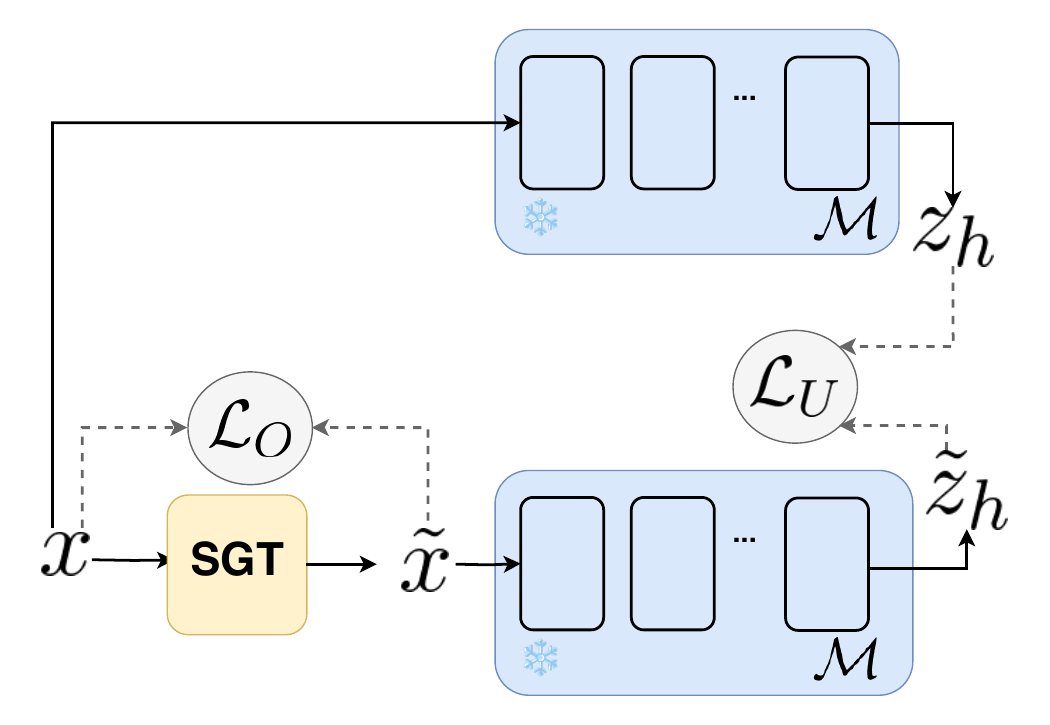}%
    \label{fig:sgt-block-train}
}
\caption{
}%
    A high-level diagram of inference, \subref{fig:sgt-block-inf}, and training, \subref{fig:sgt-block-train}, for SGT models with sequence and learnable-parameters  notation suppressed for ease of reading. Subfigure \subref{fig:sgt-block-inf}: During inference the input sequence, $x$, is used to generate a mean, $\mu(x)$, and a covariance matrix, $\Sigma(x)$. These define a multivariate Gaussian, $\mathcal{N}$, which is sampled from and added to the input to produce its obfuscations $\obfsx$. The obfuscation is then fed through the model, $\mathcal{M}$, as a usual embedding. Subfigure \subref{fig:sgt-block-train}: The SGT is trained by applying the obfuscation loss, $\LossO$, to the input and the obfuscation (dotted). To compute the utility loss, $\LossU$, the input, $x$, and its obfuscations, $\obfsx$, are passed through the frozen model to compute logits, $z_h$ and $\tilde{z}_h$, respectively, which are then compared with a cross entropy loss.
    % is fed through the model and the logits of the target model, $z_h$, are compared to the logits generated by the obfuscations (dotted), $\tilde{z}_h$. 
    In this way the SGT can learn to generate protected embeddings that maintain the utility of the target model without modifying the model at all. %
\label{fig:sgt-block-diagram}
\end{figure*}

% -------------------------------------
%   Related Works
% -------------------------------------
\subsection{Related Works} 

In this section we cover relevant related works regarding privacy technologies for LLMs and privacy techniques which are similar in formulation to the Stained Glass Transform.

\textbf{Tokenization and Masking} One of the most straightforward ways to protect a prompt is simply to replace any token which is sensitive with a placeholder token. When the sensitive token is replaced with a meaningless token unrelated to the sensitive data, the technique is known as masking (e.g replacing "Jon Doe: 555-55-5555" with "[MASK]: [MASK]"). If instead the sensitive token is replaced with an obfuscated hash which is reversible to the original token after processing, the technique is known as tokenization (e.g replacing "Jon Doe: 555-55-5555" with "[USER12345]:[VALUE12345]" prior to processing). Both of these techniques rely on applying a series of rules or possibly other language models to replace the sensitive tokens with their masked or tokenized counterparts \cite{grabler2024privacy,YAN2025100300}.
 
Tokenization and masking both can damage LLM utility when applied too liberally, and, when the sensitive data is necessary for the LLM to properly, reason and respond to the query \cite{10448234}. Moreover, they can only be applied when it is easy to identify sensitive data from non-sensitive data. This is an extremely difficult task in the case of unstructured text inputs to an LLM as pieces of data which are independently non-sensitive can become sensitive when combined together in a prompt. Due to this, tokenization and masking generally suffer from either being \emph{not enough} to provide privacy in a meaningful way or \emph{too much}, resulting in a non-trivial loss of LLM performance. 
 
\textbf{Abstraction} Another option to rules or LLM based prompt modification is abstraction \cite{dou-etal-2024-reducing,west-etal-2019-bottlesum}. This technique relies on replacing or withholding specific information in a prompt which may be sensitive. Any information which is replaced is done so with more general information that is not sensitive, but which will not dramatically affect the response of the LLM. Optionally withheld information can then be used in conjunction with the LLM response on the redacted information complete the inference \cite{siyan2025papillonprivacypreservationinternetbased}. For example in the first case, the question "What is the weather like in London, {SW1P 4GH}?" could be replaced with "What is the weather in central London?" and in the second it could be replaced with "What is the weather in London {SW1P 4GH}, San Diego {92128}, and Austin {78701}".

Abstraction can suffer from similar problems to tokenization in masking with regards to sensitive feature identification and LLM utility. It is also unclear how practical these methods are for scaling out to enterprises which may be dealing with many different types of sensitive data.
 
\textbf{Split Learning} Split learning learning aims to protect sensitive data by decomposing an existing neural network (e.g a foundational model) into two components, a client side network which is located in the trust zone of the user, and a service side component which lives in an untrusted environment \cite{10.5555/3692070.3693465,10786313,shen2023splitandprivatizeframeworklargelanguage,vepakomma2018splitlearninghealthdistributed,10.5555/3692070.3693465,evgenidis2024splitlearningcomputervision,10041745}. The output of the client side model is transformed, either by fine-tuning the client side model, or by directly applying transformations to client's output activations. Information theoretic quantities such as mutual information or distance correlation (or bounds on those quantities), can be used as losses to generate obfuscations to be transmitted to the server outside the client's trust zone \cite{10.1145/3442381.3449965,9346367}. 

This approach is straightforward to apply in non-autoregressive tasks such as image and text classification, but becomes a major engineering challenge when applied to autoregressive LLMs as significant communication is required between the client and server at every step of the autoregression. However, when the split point between the client and the server is at the embedding layer of the LLM, communication between the client and the server is not required at every step of the autoregression, allowing for efficient generation of text for transformed input embeddings. This observation is what drives the central hypothesis of the paper: \textit{it is possible to protect input embeddings to an LLM at inference time via information theoretic techniques without significantly degrading utility.}

% -------------------------------------
%   Notation
% -------------------------------------
\subsection{Notation}
For consistency in the rest of the paper, we introduce some notation in this subsection. The dataset of inputs used to train the SGT model is a random variable in ${X \in L^2(\R^d; \R^d)}$ with realizations written as $x$. The probability distribution of $X$ is denoted as $\Prob[X \in B \subset \R^d] = P_X(B)$. The SGT of the dataset is a new random variable $\obfsX$ with realizations $\obfsx$. The SGT model is a conditional probability model of the $\obfsX$ conditioned on a realization of the input, $x$, which we denote by $\Prob[\obfsX \in B | X=x] = P_{\obfsX|x}(B)$. The SGT model is trained with a utility loss, denoted $\LossU$, and an obfuscation loss, denoted $\LossO$. We use $\Model$ for the (frozen) target LLM of an SGT model.

The obfuscation loss (in part) depends on the (differential) entropy of various inputs and we will write $h(V)$ for the entropy of a random variable $V$. The conditional entropy of $V$ given another random variable $W$, is written as $h(V | W)$. When using SGTs for LLMs we work with sequences of vectors (or matrices). We write a sequence of vectors as $V_{1:T} = [V_1, \hdots, V_T]$. Finally, for a square matrix $A$, its determinant is denoted $|A|$, and the $A$-weighted $L^2$ norm is expressed as $|x|^2_A = \langle x | A | x \rangle$. A Gaussian distribution of mean $\mu$ and covariance $\Sigma$ is denoted $N(\mu,\Sigma)$, and we define $\Delta^+_d$ as the set of diagonal $d$ by $d$ positive definite matrices. The set of non-negative real numbers is denoted $\R^+$.

% % -------------------------------------
% %   The SGT
% % -------------------------------------

% -------------------------------------
%   Obfuscation Mechanisms
% -------------------------------------
\section{Stained Glass Transform: Theoretical Foundations}
\label{sec:sgt}

In this section we develop the theory of our obfuscation mechanism, the Stained Glass Transform (SGT). First we define Affine SGTs, a special class of SGTs that will be used in this paper. We then state the general definition of the SGT and provide a proposition which connects the two definitions. The general definition allows us to characterize the distribution of the full SGT of a dataset. This will serve as the cornerstone of our derivation of the information theoretic loss used to train SGT models.
\begin{definition}
    \label{def:affine-sgt}
    Given a random variable $X$ and a parameterized family of probability densities, $G(\cdot; \alpha_\theta(x)): \R^d \to \R^+$, the \textbf{Affine Stained Glass Transform} of $X$ is defined as
    % The \textbf{Affine Stained Glass Transform} of a random variable $X$ is a new random variable $\obfsX$ defined by:
    \begin{equation*}
        \obfsX = X + U
    \end{equation*}
    where $U$ is distributed according to $G(\cdot; \alpha_\theta(x))$ and $\alpha_\theta$ represents learnable parameters of a distribution (e.g. mean and covariance).
    % the (input dependent) Gaussian distribution $N(\mu_\theta(X); \Sigma_\theta(X))$. The mean $\mu_\theta: \R^d \to \R^d$ and covariance $\Sigma_\theta: \R^d \to \Delta^+_d$ are parameterized by learnable parameters $\theta$.
\end{definition}

Creating an (Affine) SGT for a random variable (i.e dataset) $X$ amounts to defining the family of probability densities. 
For LLMs the dataset consists of sequences of token embeddings from an LLM's embedding table, $\{x_i\}_{i=1}^V \subset \R^d$. 
% Most all state of the art LLMs have an embedding table $\{x_i\}_{i=1}^V \subset \R^d$, i.e. token $i$ in the vocabulary has embedding vector $x_i$. 
In this paper we choose the conditional densities to be sequence dependent Gaussian distributions. Specifically, these distributions are $N(\mu_\theta(X), \Sigma_\theta(X))$ where $%
\mu_\theta: %
\R^{d\times T} \to \R^{d \times T}%
$
and
$%
\Sigma_\theta: \R^d \to (\Delta^+_d)^T%
$ 
and where $\R^{d \times T}$ are sequences of embeddings and  $(\Delta^+_d)^T$ are sequences of diagonal positive definite matrices of length $T$. 
The sequence dependent mean and variance estimator functions are modeled as transformers to ensures that the SGT models are a function of the \emph{sequence} of the embeddings rather than treating embeddings as independent of each other. Figure~\ref{fig:sgt-block-inf} gives a description of inference.

\textbf{Inference with SGT models.} Current state of the art LLMs  take as input sequences of embeddings from a fixed embedding table. The affine structure of the SGT allows us to easily generate obfuscations of sequences of embeddings. 
Given an input sequence $x_{1:T} \in \mathbb{R}^{d\times T}$ we generate an 
obfuscation by computing the means $\mu_\theta(x_{1:T})_{1:T}$ and variances 
$\Sigma_\theta(x_{1:T})_{1:T}$. Next, we sample $T$ independent multivariate standard normal vectors, $u_{1:T} \sim \mathcal{N}(0_d, I_d)$, creating the SGT:
\begin{equation}   
\label{eq:stg-llms}
    \Tilde{x}_{1:T} %
        = x_{1:T}
        + \mu_{\theta}(
            x_{1:T}
        )_{1:T}
        + \Sigma^{1/2}_{\theta}(
            x_{1:T}
        )_{1:T}\  \ 
        u_{1:T}.
\end{equation} 
%

% The means and variances are modeled using the initial layers from the target model with an additional learnable head. 
\textbf{Preparing for training.} Equation \eqref{eq:stg-llms} shows that obfuscation with an SGT is entirely defined by the conditional probability $P_{\obfsX|X=x}$. However, in order to learn the parameters of the model ($\mu_\theta$ and $\Sigma_\theta$), we need to understand how the SGT itself is distributed. This is most easily done using the general definition of the SGT:

\begin{definition}
    \label{def:sgt}
    Given a random variable $X$ on $\R^d$, a \textbf{Stained Glass Transform} of $X$ parameterized by density family  
    $G(\cdot; \alpha(x)): \R^d \to \R^+$, 
    is defined as a random variable $\obfsX$ whose (regular) conditional probability satisfies:
    \begin{equation*}
        % \label{eq:stg-cond}
        % P_{Y|X=x}(B; \theta) = \int_B \rho(y; x, \theta) dy.
        P_{\obfsX|X=x}(B; \theta) = \int_B G(y ; \alpha_\theta(x)) dy.
    \end{equation*}
\end{definition}

The following proposition provides a description of the conditional densities of Affine SGTs.

\begin{proposition}
\label{prop:affine-sgt}
      Let $\obfsX$ be an Affine SGT of $X$ with $\obfsX= X + U$ where $U \sim G(\cdot; \alpha)$, the the conditional probability of $\obfsX$ given $X=x$ is:
      \begin{align*}
        P_{\obfsX|X=x}(B)
            &= \int_B G(y - x; \alpha(x))dy
    \end{align*}
\end{proposition}

\begin{proof}
    Assume that $U$ has conditional density $G(u; \alpha(x))$ and consider the random variable $\obfsX = X + U$. Then
    \begin{align*}
        P_{\obfsX|X=x}(B)
            &= \Probof{\obfsX \in B | X = x}  \\
            &= \Probof{\obfsX - X \in B - x | X = x} \\
            &= \Probof{U \in B - x | X = x}
    \end{align*}
    Which in integral form becomes $ \int_{B - x} G(u ; \alpha(x)) du = \int_B G(y - x; \alpha(x))dy$.
\end{proof}

Proposition \ref{prop:affine-sgt} characterizes the distribution of the SGT, $\obfsX$, over a dataset. The empirical distribution of our dataset is given by $P_X(x) = \sum \omega_i \delta(x - x_i)$, where $\delta(x-\cdot)$ is a point mass density. Proposition \ref{prop:affine-sgt} then implies the following corollary.

\begin{corollary}
    \label{cor:affine-sgt-gmm}
    Let $\obfsX$ be an Affine SGT of $X$ of the form $\obfsX=X+U$ with $U \sim N(\mu_\theta(X); \Sigma_\theta(X)$, then $Y$ is distributed as a Gaussian Mixture Model:
    \begin{equation}
        \label{eq:sgt-gmm}
        P_{\obfsX}(z) = \sum_{x_i \in X} \omega_i N(z; \ \mu_\theta(x_i) + x_i; \ \Sigma_\theta(x_i))
    \end{equation}
    where $\omega_i = P_X(x_i)$.
\end{corollary}

Knowing the precise form of the obfuscation allows us to derive information theoretic relationships between our the SGT and its input dataset. In particular it we can now rigorously incorporate the mutual information between the input dataset and the obfuscation into our learning objective.
% For simplicity we often write $N(z; x)$ for $N(z;\mu_\theta(x_i) + x_i;\Sigma_\theta(x_i))$.
% %---------------------------------
% %   Learning Affine-SGTs
% %---------------------------------

% -------------------------------------
%   The Learning Problem
% -------------------------------------
\section{Obfuscation Learning Problem}
\label{sec:learning-prob}
In this section we explain the loss function used to train the SGT for LLMs. Given a target model $\Model$ and a dataset 
% $X \in L^2(\Omega_1; \R^d)$ 
$X$ 
we want the obfuscation
% $\obfs{X} \in L^2(\Omega_2; \R^d)$ which satisfies:
$\obfs{X}$ to satisfy:

\begin{enumerate}
    \item \textbf{Utility:} The obfuscation should maintain similar performance with respect to $\Model$.
    \item \textbf{Privacy:} The obfuscation should not be reconstructible.
\end{enumerate}

To achieve both goals we take a multi-objective learning approach with two classes of losses: a utility loss, $\LossU$, that penalizes the obfuscation for degrading the performance of $\Model$ and a obfuscation loss, $\Loss_O$, which encourages the obfuscation to be difficult to reconstruct.

For our utility loss we use a self-supervised loss function at the logits layer of the frozen target LLM, $\Model$ between the logits arising from the SGT and untransformed input embeddings. Specifically, if $z_h$ and $\obfs{z}_h$ are the logits of the target model for inputs $x$ and $\obfsx$, respectively, then the utility loss is: $\LossU(x, \obfsx) = \text{d}(z_h, \obfs{z_h})$
%
% \begin{equation}
%     \label{eq:distil-loss}
%     \LossU(x, \obfsx) = \text{d}(z_h, \obfs{z_h})
% \end{equation}
%
where $\text{d}$ is some metric on $\R^d$. In practice we choose $\LossU$ to be cross entropy loss between the probabilities generated by the logits of the SGT and untransformed input embeddings. The obfuscation loss, $\LossO$, is to reward increasing the difficulty in reconstructing $X$ from $\obfs{X}$. 
% Motivation and details are presented in Sections \ref{sec:computing_mi_loss} to \ref{sec:other_obfucsation_loss_components}. 
The full loss function is a sum of the utility and obfuscation losses:
\begin{equation}
    \label{eq:basic-loss}
    \Loss = \LossU + \LossO.
\end{equation}
In practice the obfuscation loss is made up of three subcomponents (detailed in Section \ref{sec:other_obfucsation_loss_components}).

{Figure~\ref{fig:sgt-block-train} shows the procedure for training an SGT model for a fixed target LLM, $\Model$.} Broadly, we freeze the weights of $\Model$ during training and only update the SGT model's weights. 
% The specific losses are discussed below in Section \ref{sec:learning-prob} but we give a brief overview here. 
The input, $x_{[1:T]}$, is compared against its SGT realization, $\obfsx_{[1:T]}$, via the obfuscation loss. Then both the input embedding and its SGT are fed through $\Model$. The resulting logits are compared with cross entropy.
% to encourage the SGT to preserve the performance of the LLM. 
In this way an SGT model can be trained without modifying the target LLM.

%----------------------------------------------------
%       PAC Privacy
%----------------------------------------------------
\subsection{Mutual information and PAC Privacy bounds}
\label{sec:PAC Privacy}
In this subsection we describe how we use mutual information in the obfuscation loss $\LossO$ to provide information theoretic protection of embeddings. We begin by motivating its consideration.

Mutual information has been widely used as a measure of privacy \cite{duan2024reimagining, mai2023split, dou-etal-2024-reducing, makhodoumi_bottleneck_2014, alemi_dvib_2016}. Moreover, in \cite{xiao2024formal} the authors develop a provable-approximately-correct theory of privacy (PAC-Privacy) which gives theoretical guarantees that bound the probability an adversary can reconstruct an input from its obfuscation. The key result is that, given an input $X$ and obfuscation mechanism $\obfsX$, then the reconstruction advantage satisfies the bound
\begin{equation}
    \label{eq:pac-bound}
    \Delta_{KL}\delta \leq MI(X; \obfs{X}),
\end{equation}
where the $\Delta_{KL}\delta = \delta \log(\frac{\delta}{\delta_0}) + (1 - \delta) \log(\frac{1 - \delta}{1 - \delta_0})$, and is the Kullback-Leiber Divergence between the a-priori success rate, $1-\delta_0$, of reconstruction (i.e. the likelihood an adversary can reconstruct an input knowing only the data distribution) and the the posterior advantage, $1-\delta$, gained by observing the obfuscation. This result demonstrates that mutual information plays a key role in privacy protection.

However, Bound \eqref{eq:pac-bound} was used by the authors to construct a specific obfuscation mechanism that was tractable to compute. Our situation is attempting to measure the protection of more general obfuscation mechanisms and, as was noted by the authors in \cite{pacalg:2024}, the bound shown in Equation~\eqref{eq:pac-bound} is not tight and in practice stronger privacy is observed. 

This implies that the precise relationship between reconstruction difficulty and mutual information is more complex than in Equation \eqref{eq:pac-bound}.
% there are additional facets to privacy protection beyond mutual information. 
We believe that future work to improve the bound in Equation \eqref{eq:pac-bound}, even if only for homogeneous mixture models, would allow for a better understanding of the privacy offered by common noise mechanisms beyond the worst-case guarantees of differential privacy.

% In Section~\ref{sec:experiments}, we report the mutual information (\textbf{MI}) of our models approximated via a Monte-Carlo approximation on the testing dataset. The mutual information is computed at the feature level rather than the whole embedding to account for the high-dimensional embeddings of the target model. 

% By choosing an a-priori reconstruction success rate, $1-\delta_0$, one can invert Equation \eqref{eq:pac-bound} to obtain an explicit bound on the reconstruction probability $\delta$. We report these bounds (PAC-Adv) in Table~\ref{tab:loss-ablation} and approximate the a-priori success as $1/N$, where $N$ is the number of token embeddings. Importantly, any mutual information value above $\log(N)$ provides an upper bound on reconstruction success of more than 1 and so falls out of the scope of Equation \ref{eq:pac-bound}. Since our mutual information is computed at the feature level these bounds represent the robustness of reconstruction of individual features of token embeddings.

% -------------------------------------
%   MI Loss
% -------------------------------------
\section{Controlling Mutual Information}
\label{sec:computing_mi_loss}
The preceeding section provides a theoretical justification for using mutual information in our obfuscation loss. In this section we describe mutual information and shortcomings of common approximation method. We then use Corollary \ref{cor:affine-sgt-gmm} to derive a mon
% In this subsection we introduce the information theoretic tools we use in the obfuscation loss $\LossO$ to provide information theoretic protection of embeddings.

% Mutual information has been widely used as a measure of privacy \cite{duan2024reimagining, mai2023split, dou-etal-2024-reducing, makhodoumi_bottleneck_2014, alemi_dvib_2016}. Moreover, the recent work in PAC-Privacy by \cite{pacalg:2024, xiao2024formal} provides theoretical guarantees on bounding reconstruction probability by mutual information. This motivates developing a loss to directly control the mutual information of our obfuscations.

% Recall that the PAC-Privacy framework provides rigorous bounds on the reconstruction success via mutual information  . 
The mutual information between a dataset, $X$, and its obfuscation, $\obfsX$, can be written:
\begin{equation}
    \label{eq:mutual-info}
    MI(X; \obfsX) = h(\obfsX) - h(\obfsX|X)
\end{equation}
where $h(Z)$ and $h(Z|X)$ are the differential entropy of $Z$ and of $Z$ conditional on $X$ respectively. Incorporating mutual information into our obfuscation loss requires approximating the entropy of the obfuscated dataset as well as the conditional entropy of the obfuscations and its input. For SGTs, Proposition \ref{prop:affine-sgt} implies $h(\obfsX | X) = \Expsub{X}{h(\ N(y; X)\ )}$ is the expected entropy of the learned family of Gaussians, which is straight forward to compute.

The entropy of the obfuscation itself, $h(\obfsX)$, requires more care. Corollary \ref{cor:affine-sgt-gmm} implies the obfuscations are distributed according to a high dimensional Gaussian Mixture Model (GMM). The entropy of a GMM does not have a closed form expression, so common approaches to compute it rely on bounds and approximations \cite{huber2008entropy, eskenazis2018gaussian, kolchinsky2017estimating}.

% -------------------------------------
%   Issues With Mixture Entropy Bounds
% -------------------------------------
\subsection{Avoiding Common Mixture Entropy Bounds}
\label{sec:improving_entropy_bounds}
Before providing our approximation method for the mutual information loss, we first show why care must be taken it. We are not interested in just a close numerical approximation of the entropy, but rather we need an approximation which provides an adequate signal for the obfuscation mechanism to reduce the mutual information.

To demonstrate potential shortcomings of naive implementations, note that for a GMM the conditional entropy term in Equation \eqref{eq:mutual-info} is the expected value of: $h(N(y; \mu(x), \Sigma(x)) \approx -\log(|\Sigma(x)|)$. The remaining term is the entropy of the GMM, $\obfsX$. A common upper bound for a Gaussian mixture model's entropy is to use the entropy of the components themselves \cite{gradEntropyEst-Schraudolph-2004, SHWARTZ20051045, empircalEntropyViola-1998}. Specifically,
\begin{equation}
    \label{eq:gmm-entropy-ub}
    h(\obfsx) \leq \sum_i \omega_i h_y( N(y;  \mu(x_i), \Sigma(x_i)) )
\end{equation}
where $h_y$ denotes entropy computed with respect to the $y$ variable.

Equation \eqref{eq:gmm-entropy-ub} provides a useful bound and can in some cases be a strong approximation of the entropy itself \cite{kolchinsky2017estimating}.
Unfortunately, it has a shortcoming that prevents it from providing a signal to minimize mutual information: there is no penalty for increasing information \emph{between} the mixture components. This is because conditional entropy only depends on the variance of the components 
(i.e. $\log(|
\Sigma|)$) 
and does not depend on the means. Using Equation \eqref{eq:gmm-entropy-ub} for our mutual information loss would result in a signal in which increasing the variance of the Gaussian components is the \emph{only} way to decrease the mutual information. The resulting suboptimal performance is demonstrated empirically in Section \ref{sec:experiments}.

% -------------------------------------
%   MI Loss MBMC
% -------------------------------------
\subsection{Mutual Information Loss with Minibatch Monte Carlo}

% [H] because IEEE does not allow algorithms to float
% https://tex.stackexchange.com/questions/219816/algorithm-in-ieee-format
\begin{algorithm}[H]
    \caption{Minibatch Mixture Entropy Approximation}
    \label{alg:entropy-approx}
    \begin{algorithmic}[1]
        \REQUIRE  Obfuscation Sampler $P(\obfsX \mid x)$, Batch sampler $D$, Density model $G(z ; x)$, Batch size $B$
        \STATE Sample two independent batches: $X_b \sim D$, $X_b' \sim D$
        \STATE Initialize entropy estimate: $\mathcal{H} \gets 0$
        \FOR {$i = 1$ to $B$}
            \STATE $x^1 \gets X_b[i]$
            \STATE Sample $\obfsx^1 \sim P(\obfsX | x^1)$
            \STATE $x^2 \gets X_b'[i]$ \quad \# Cross-sampling for mixture entropy estimation
            \STATE $\mathcal{H} \mathrel{+}= -\log G(\obfsx^1 ; x^2)$
        \ENDFOR
        \STATE \textbf{return} $\mathcal{H} / B$
    \end{algorithmic}
\end{algorithm}

We now present our mutual information based obfuscation loss for training SGTs. 
Though we only experiment with multivariate Gaussians for our SGT density families, the SGT is defined for any family of (regular) conditional distributions (Definition \ref{def:sgt}). For this reason, we present Algorithm \ref{alg:entropy-approx} for a general family of conditional densities $G(z; x)$. The  mutual information can then be written as:
\begin{align}
    &MI(X; \obfsX)
        = \label{eq:mi stacked exp}\\
        &\quad 
        \Expsub{z\sim \obfsX}{
                -\log\left(
                    \Expsub{X}{G(z; x)}
                \right)
            }
        -\Expsub{X}{
            h_z(G(z; x))
        }.
        \nonumber
\end{align}
If we require that the  $h_z$ term have an analytic expression for the family of densities (e.g. Gaussian, Laplacian, Exponential, Cauchy), then it can be computed directly across minibatches. To approximate the remaining expectation in Equation \eqref{eq:mi stacked exp} we use a minibatch Monte-Carlo approximation of the entropy, detailed in Algorithm \ref{alg:entropy-approx}.

Broadly, we sample a batch of inputs, 
 $\{x_i\}_{i=1}^B$, 
and compute their obfuscations 
 $\{\obfsx_i\}_{i=1}^B$. 
Then we \emph{independently} sample another batch of inputs, $\{x_i'\}_{i=1}^B$,
and compute the average negative log probability between the samples. The two entropy terms are then subtracted and averaged across the batch.

Note in Algorithm \ref{alg:entropy-approx} the negative log probability term, $-\log G(\obfsx^1 ; x^2)$. This term is the negative log probability of an obfuscation with respect to an input that did not generate the obfuscation. \emph{Thus, using Algorithm \ref{alg:entropy-approx} ensures that the obfuscation loss provides a signal on how mixing of components contributes to the mutual information of the obfuscation.}

Since our density family are Gaussians there exists a closed form for the Monte-Carlo approximation of the mutual information loss between two batches $X_b$, $X_b'$, and the obfuscation $\obfsx_b$ of $X_b$. 

\begin{proposition}
    \label{prop:gasgt-mi-loss}
    Let $X_b$ be a batch sample of size, $B$, from our dataset $X$ with SGT $\obfsX_b$, and let $X_b'$ be an independently sampled batch. Further, denote $\Sigma_k = \Sigma(x_k)$ and $\mu_k = \mu(x_k)$ the estimated covariance and mean of the SGT model, respectively. 
    Then the Monte-Carlo approximation of the obfuscation loss in Algorithm \ref{alg:entropy-approx} is:

    \begin{align}
        \LossMI(X_b, &\obfsX_b; X_b') \ = \nonumber\\
        &\frac{1}{B}\sum_{\ell', i} \log(
            |\Sigma_i^{-1}\Sigma_{\ell'}|
            )
        + \|
            \obfsx_i - x_{\ell'} - \mu_{\ell'} 
        \|_{\Sigma_{\ell'}^{-1}}^2
        \label{eq:sgt-llm-noise-loss}
    \end{align}
where $x_i$ is an element of $X_b$, $\obfsx_i$ is its corresponding SGT, and $x_{\ell'}$ is an element of $X_b'$ (Terms constant in $\theta$ have been dropped).
\end{proposition}

The mutual information loss in Equation \eqref{eq:sgt-llm-noise-loss} can be decomposed into three pieces. The first is $\log(|\Sigma_i^{-1}|)$ which is the entropy of the components of the GMM. This term encourages higher variance in each component and is the loss that we arrive at from Bound \eqref{eq:gmm-entropy-ub}. When considering the full mutual information however; the component-wise entropy is mediated by the entropy of the sampled mixture components 
( the $\log(|\Sigma_i^{-1}\Sigma_{\ell'}|)$ term). This encourages the  covariances to be more similar to one another. This avoids having an individual component with much larger covariance than other components standing out, which would leak information. Finally, the loss includes a Mahalanobis distance term between components of the mixture, penalizing individual components whose distribution has drifted from the centers of the rest of the mixture. In this way reducing the mutual information loss, results in reducing the identifiably of any given component.

% \paragraph{Geometric Catalyst Losses}
% \label{sec:Cat Losses}
% We found that the mutual information loss can be slow to converge and requires careful hyperparameter tuning. To address this we introduce an auxiliary Geometric Catalyst Loss whose goal is to encourage the SGT to learn early on to move embeddings. Geometric Catalysts should be decreasing functions of a distance measure (euclidean norm, cosine similarity, etc) between the obfuscation and its clean embedding. In this work we use an inverse power law of $L_C(\obfs{X}, X) = \eta / \|\obfs{X} - X\|_2$ as the Geometric Catalyst. This catalyst penalizes the model for having obfuscations very near the input. We choose $\eta$, referred to as the characteristic length, to be the fifth percentile of pairwise distances between tokens in the model's embedding table. The fifth percentile worked well across our experiments and optimal selection of this parameter is saved for future work.

%-------------------------------------------
%       Other Obfuscation Loss Components
%-------------------------------------------
\subsection{Other Obfuscation Loss Components}
\label{sec:other_obfucsation_loss_components}
In this subsection we describe the two other components which contribute to the obfuscation loss function $\LossO$.
In practice the mutual information loss, $\LossMI$, presents a difficult optimization problem and can be slow to converge or sensitive to hyperparameters. By adding additional terms to the obfuscation loss function $\LossO$ we are able to both speed up convergence of the SGT and produce SGTs with higher quality obfuscations.
% The mutual information loss $\LossMI$ can suffer from slow convergence and lower than desired obfuscation scores when used in isolation.

\textbf{Absolute Cosine Similarity (\sgtabscos)} During training of the mutual information loss it was observed that the cosine similarity between the input embeddings, $X_b$, and their SGT, $\obfsX_b$, slowly converged to zero. Geometrically this indicates the states of lowest mutual information which still maintain utility are orthogonal to the original input embeddings. We choose to accelerate this behavior by adding a loss component $\LossACS$ which computes the absolute value of the cosine similarity between a batch of input embeddings $X_b$ and their SGT $\obfsX_b$:
\begin{equation}
\label{eq:absolute-cosine-loss}
    \LossACS(X_b, \obfsX_b) = \left| \frac{X_b \cdot \obfsX_b}{||X_b||||\obfsX_b||} \right| 
\end{equation}

\textbf{Median Norm Penalty (\sgtnormpen)} During training with mutual information loss, the transformed input embeddings norm tended to drift away from the original distribution of the embedding, slowing convergence. To combat this, we include a loss which penalizes the norm of the mean component the SGT:
% of the input embedding from being far from the median of the norms of the constituent vectors comprising the input embedding table:
% 
\begin{equation}
\label{eq:median_norm_penalty}
    \LossMNP(X_b) = 
        |\ 
            \| X_b + \mu_\theta(X_b) \|  - T \ 
        | ,
\end{equation}
where $\mu_\theta(X_b)$ is the mean of the Gaussian distribution comprising the SGT. We set $T$ to be the median norm of the embedding vectors (i.e. $T = \text{Median}(\{||e|| : e \in E \})$ where $E$ is the embeddings table of $\Model$). The total obfuscation loss is then the weighted sum of its constituent components:

\begin{equation}
\label{eq:full-loss}
    \LossO = \alpha_1\LossMI + \alpha_2\LossACS +  \alpha_3\LossMNP,
\end{equation}
where each $\alpha_i \in \R^+$.

% -------------------------------------
%   Privacy Metrics
% -------------------------------------
\subsection{Demonstrative Obfuscation Losses}
\label{sec:demo-loss}

In this section we describe two obfuscation loss functions that we experimented with to demonstrate useful properties of the final obfuscation loss, $\LossO$. The first is the component wise Gaussian entropy (\textbf{\sgtlog}) from Bound \eqref{eq:gmm-entropy-ub}:
\begin{equation}
\label{eq:loss-log}
\LossCGE(\obfsX_b) = 
    \log\left(
        -| \Sigma_\theta(X_b) |^{-1}
    \right).
\end{equation}
The second is a cosine-similarity penalty (\textbf{\sgtcos}) which does not encourage orthogonality, but rather antipodality:
\begin{equation}
\label{eq:loss-cos}
    \LossBCS(X_b, \obfsX_b) = 
        \frac{X_b \cdot \obfsX_b}{||X_b||||\obfsX_b||}.
\end{equation}

Both losses encourage differences between inputs and obfuscations but as we will show in the results they provide in weak or superficial protection of data.
% -------------------------------------
%   Privacy Metrics
% -------------------------------------
\section{Privacy Metrics}
\label{sec:metrics}

% \subsection{Reconstruction Attack Robustness}
% \label{sec:metrics priv}
%--------------------
%
%   ReconRank Hist: CosSim
%
%--------------------

\begin{figure}
    \centering
    \includegraphics[width=0.75\linewidth]{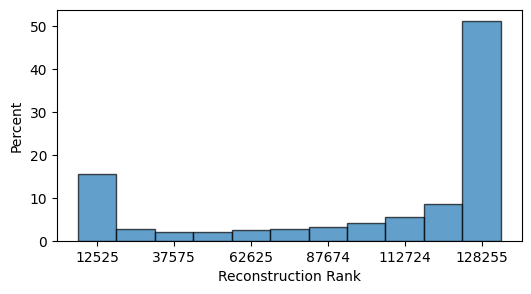}
    \caption{A histogram of  reconstruction rank of obfuscations from an SGT trained with cosine-similarity obfuscation loss as opposed to considering mutual information which was discussed in Section~\ref{sec:learning-prob}. The skew towards the farthest ranks demonstrates a failure of this model not captured by the common nearest neighbor reconstructions attack.}
    \label{fig:recrank-hist-cossim}
\end{figure}

\emph{A primary measure of an obfuscations strength is its resilience to reconstruction attacks.} Many reconstruction attacks are based on the nearest clean neighbors in the embedding table to the obfuscation \cite{mai2023split, ZHANG2025104293, li2023privacy}. In this section we define the nearest neighbor attacks, language aware attacks, and then provide empirical evidence that the distance of an obfuscation to the other embeddings in the LLM's embedding table provides an attack vector missed by nearest neighbors. 
% \textcolor{red}{We show that SGT's trained to minimize mutual information provide resilience to such attacks.}
% show that empirically evaluating the privacy level provided by obfuscation techniques should more robustly measure retrieving information from obfuscated data beyond nearest clean neighbors. 

%------------------------------------
%       Nearest Neighbor Recon
%------------------------------------
\subsection{Nearest/$\text{N}^{\text{th}}$ Neighbor Reconstruction}

Define the reconstruction rank of an obfuscation $\obfs{X}$ of clean embedding $X_i$ to be 
\begin{align*}
 R(\obfs{X}) = 
     &\rank_i\left( \ 
        \{
            \|\obfs{X} - X_k\|_2 \ : \ k \in \{1, \hdots, |V| \}
        \}\right). %\\
        % &SGT(X_i) = \obfsX
\end{align*}
That is, $R(\obfsX)=r$ means that the clean token embedding, $X$, which generated $\obfsX$ is the $r^{\text{th}}$ closest token embedding to $\obfsX$.

The relative distance of an obfuscation to each clean token carries valuable information. In Figure \ref{fig:recrank-hist-cossim}  we compute the reconstruction rank of obfuscations across the Alpaca dataset and plot its histogram. This specific SGT was trained with the \sgtcos loss (see Equation \eqref{eq:loss-cos}) rather than obfuscation loss $\LossO$ in Equation \eqref{eq:full-loss}. With this basic cosine-similarity loss, we see a bimodal behavior in the rank histogram. There is a small peak at the nearest neighbors followed by a larger peak at the \emph{farthest} neighbors, implying that:

\emph{Reconstructions based on the nearest neighbors would perform worse than those which considered the farthest ones\footnote{Note this assumes additional greybox information about the SGT, e.g. some knowledge of the rank histogram or ability to run an SGT on a small dataset.}.
}

This motivates two reconstruction attacks for a more robust evaluation obfuscations:
\begin{enumerate}
    \item \textbf{NN}: The nearest neighbor reconstruction of an obfuscation $\obfs{X}$ is $X_{i^*}$ where $i^*$ satisfies $R(\obfs{X}, i^*) = 1$. 
    \item \textbf{MRP}: Given a maximum rank probability $r$ over a dataset $D$, the max-rank-probability reconstruction (MRP) of an obfuscation $\obfs{X}$ is $X_{i^*}$ where $i^*$ satisfies $R(\obfs{X}, i^*) = r$. That is, it is the $r^{th}$ farthest clean neighbor to the obfuscation. 
\end{enumerate}

\subsection{Language Aware Reconstruction}
The state-of-the-art for reconstruction of input text from transformed embeddings is a language aware attack known as BeamClean \cite{kale2025beamcleanlanguageawareembedding}. This technique jointly estimates the noise parameters of the transformed embeddings and decodes token sequences by integrating a language-model prior.

\subsection{Transformation Metrics}

In this section we describe metrics based on how well an obfuscation changes a token embedding. We begin with metrics based on reconstruction attacks. The general robustness of obfuscations against these attacks is measured by the failure rate of the attacks:
\begin{itemize}
    \item \textbf{NN-FR}: The failure rate of a nearest neighbor reconstruction attack.
    \item \textbf{MRP-FR}: The failure rate of a max rank probability reconstruction attack.
    \item \textbf{BeamClean-FR}: The failure rate of a BeamClean attack.
\end{itemize}

% Beyond the reconstructions described in the previous subsection, fully quantifying the additional information in the rank histogram can be measured by its entropy, as it can be interpreted as measuring the vulnerability to any attack on the distribution of the rank histogram. To this end, we report the entropy of each SGT's rank histogram normalized by a uniform (i.e. maximally entropic) distribution (\textbf{Hist. Entropy}). However, knowing that an SGT has low entropy only tells us that there is information that may be exploited with no obvious way how.
Based on our empirical observations we see the peak of the rank histograms are at the nearest and farthest ranks (Figure \ref{fig:recrank-hist-cossim}) across models. For this reason we report the following metrics.

(\textbf{TTR-k}) The Token Transformation Rate at $k$  over a dataset of obfuscated tokens $D$ is
\begin{equation*}
    \text{TTR-k} = 100 \times
        \frac{
            | \{
            \obfs{X} \in D \ : \  k < R(\obfs{X}) \  \}|
        }{|D|}
\end{equation*}
i.e. the percent of obfuscated tokens whose clean embedding is not in its set of k-nearest-neighbors. Concretely, the TTR-1 metric is exactly the familiar metric of nearest clean token transformation rate \cite{mai2023split, ZHANG2025104293, li2023privacy}. 

\textbf{(SymTTR-k)} The symmetric TTR at $k$  over a dataset of obfuscated tokens $D$ is
\begin{equation*}
     \text{SymTTR-k} = 100 \times
        \frac{
            | \{
            \obfs{X} \in D \ : \  k < R(\obfs{X}) < |V| - k   \}|
        }{|D|}
\end{equation*}
i.e. the percent of obfuscated tokens whose clean embedding is not in the set of k-nearest or k-farthest neighbors. 

% Unlike entropy, the 
% TTR-k and SymTTR-k metrics have a clear geometric meaning. They quantify the accumulation of the rank histogram at the nearest and farthest tokens to the obfuscated embedding. 
The TTR-k and SymTTR-k metrics extend upon this nearest neighbors based metric to help understand not only whether the transformed embedding is mapping to a transformed token, but how the transformed embeddings are being distributed with respect to the embeddings which appear in the LLM's embedding table geometrically.

Another approach for measuring the privacy leakage of reconstructions of transformed embeddings are the PII accuracy (\textbf{PII}) metrics developed in PAPILLON \cite{siyan2025papillonprivacypreservationinternetbased}. This technique evaluates the privacy of reconstructed text, rather than token level reconstruction, embeddings by using a LLM-as-a-judge to determine whether the reconstructions contain the private attributes of the original plaintext and ensures that the reconstructions are not transforming the text in a way in which the sensitive data can be easily recovered. We report the percent of PII-Ratio in the input text which was recovered by a reconstruction. Given an input text $T$ and reconstructed text $T^*$

\begin{equation*}
    \text{PII-Ratio}(T, T^*) = \frac{
        \text{Percent of $T^*$ which is PII }
    }{
    \text{Percent of $T$ which is PII }
    }
\end{equation*}

% \textcolor{red}{\subsection{Papillon}Should we have it as its own subsection??} Another approach for measuring the privacy leakage of reconstructions of transformed embeddings is to use the metrics developed in \cite{siyan2025papillonprivacypreservationinternetbased}. This technique evaluates the privacy of reconstructed text embeddings by using a LLM-as-a-judge to determine whether the reconstructions contain the private attributes of the original plaintext and ensures that the reconstructions are not transforming the text in a way in which the sensitive data can be easily recovered. 

\subsection{Information Theoretic Metrics}

In this section we describe metrics for determining the amount of information leaked by an obfuscation. These metrics do not have direct attacks associated with them, but minimizing them increases an obfuscation's resilience across attack types.

The TTR and SymTTR metrics were \emph{a specific} way of measuring the exploitability of the rank histogram but are not exhaustive. Fully quantifying the additional information in the rank histogram can be measured by its entropy, as it can be interpreted as measuring the vulnerability to any attack on the distribution of the rank histogram. To this end, we report the entropy of each SGT's rank histogram normalized by a uniform (i.e. maximally entropic) distribution:
\begin{equation*}
    \text{Hist. Entropy} = H[R(\obfsX)] / \log(N)
\end{equation*}
where $N$ is the number of clean embedding vectors.

Additionally, we report the mutual information (\textbf{MI}) of our models approximated via a Monte-Carlo approximation on the testing dataset. The mutual information is computed at the feature level rather than the whole embedding to account for the high-dimensional embeddings of the target model. 

By choosing an a-priori reconstruction success rate, $1-\delta_0$, one can invert Equation \eqref{eq:pac-bound} to obtain an explicit bound on the reconstruction probability $\delta$. We report these bounds (\textbf{PAC-Adv}) in Table~\ref{tab:loss-ablation} and approximate the a-priori success as $1/N$, where $N$ is the number of token embeddings. Importantly, any mutual information value above $\log(N)$ provides an upper bound on reconstruction success of more than 100\% and so falls out of the scope of Equation \ref{eq:pac-bound}. Since our mutual information is computed at the feature level these bounds represent the robustness of reconstruction of individual features of token embeddings.

%--------------------------------------
%   EXPERIMENTS
%--------------------------------------
\section{Experiments}
\label{sec:experiments}

The loss ablation studies use the Llama 3.2 1B Instruct as the target model \cite{grattafiori2024llama3herdmodels} with the SGT models trained on the OpenOrca dataset \cite{OpenOrca}. Due to the number of losses tested and the need for hyperparameter tuning we train these models on 15\% of the dataset and select the best performer based on SymTTR. Obfuscation quality is evaluated on the Alpaca dataset \cite{mukherjee2023orcaprogressivelearningcomplex, alpaca}. To measure the utility of SGTs we compare a model using SGT's obfuscated embeddings with the target model on several LLM benchmarks (Arc Challenge, Arc Easy, Openbook QA, PIQA, TruthfulQA, IFEval,  MMLU 0-shot, and Hellaswag 10-shot \cite{arc2018, OpenBookQA2018, seo2018phrase, truthfulQA2021, mmlu2020, zellers2019hellaswag}). We then report the average performance difference from the target model across all benchmarks (Utility). Models were each trained on a single Nvidia A100s 80GB for roughly six hours.

Additionally, we perform evaluation on the PUPA dataset \cite{siyan2024papillon}. On this dataset we compute the NN and BeamClean failure rates. Additionally, we use the PAPILLON framework \cite{siyan2024papillon} for computing PII contained in the original text as well as the reconstructions and report the ratio (PII-Ratio). Due to the compute required for BeamClean only a subset of the SGTs trained for Llama-1B are evaluated in this way.
\subsection{Model Comparisons}

\textbf{Obfuscation Losses.} To understand the impact of the different constituent losses in \eqref{eq:full-loss} we train SGTs using only the mutual information loss (\sgtmi), only the absolute cosine loss (\sgtabscos), and the combination of the two (\sgtmiabscos). Additionally, we compare with the demonstrative losses in Section \ref{sec:demo-loss} (i.e. \sgtcos and \sgtlog).

\textbf{Constant Noise.} We experiment with constant noise obfuscations (\constnoise). For each embedding in a sequence we add a constant Gaussian and vary the strength of the noise via its variance.These are analogous to local Differential Privacy (DP) \cite{dwork_2006} and so we report the $\epsilon$-DP value for the standard deviation of the Gaussian mechanism \cite{dwork2014_dp}.

\textbf{Large Models Fully Trained.} To verify that the SGT generalize to true Large Language Models, we apply the \sgtfull loss to train SGT models for: Llama-3.3-70B, Qwen3-32B \cite{qwen3technicalreport}, and the Llama-70B distilled DeepSeek-R1-70B \cite{deepseekai2025}. These models were trained for two epochs on the full OpenOrca. Results are reported a validation subset of OpenOrca (up to half a percent of the dataset depending on number of nodes available at the time). All large models were trained with the full \sgtfull \ loss.

These larger models were trained up to 64 nodes of 8 Nvidia A100 80GB GPUs (hardware issues led to use of 32 nodes for some training restarts). Models were trained for up to 2 days depending on hardware configurations and model context lengths. Models were trained using FSDP2 x Tensor-Parallelism \cite{liang2025torchtitan}.

%-------------------------------------------------
%       RESULTS
%-------------------------------------------------
\subsection{Results}

\begin{figure}[!t]
    \centering
    \includegraphics[width=0.9\linewidth]{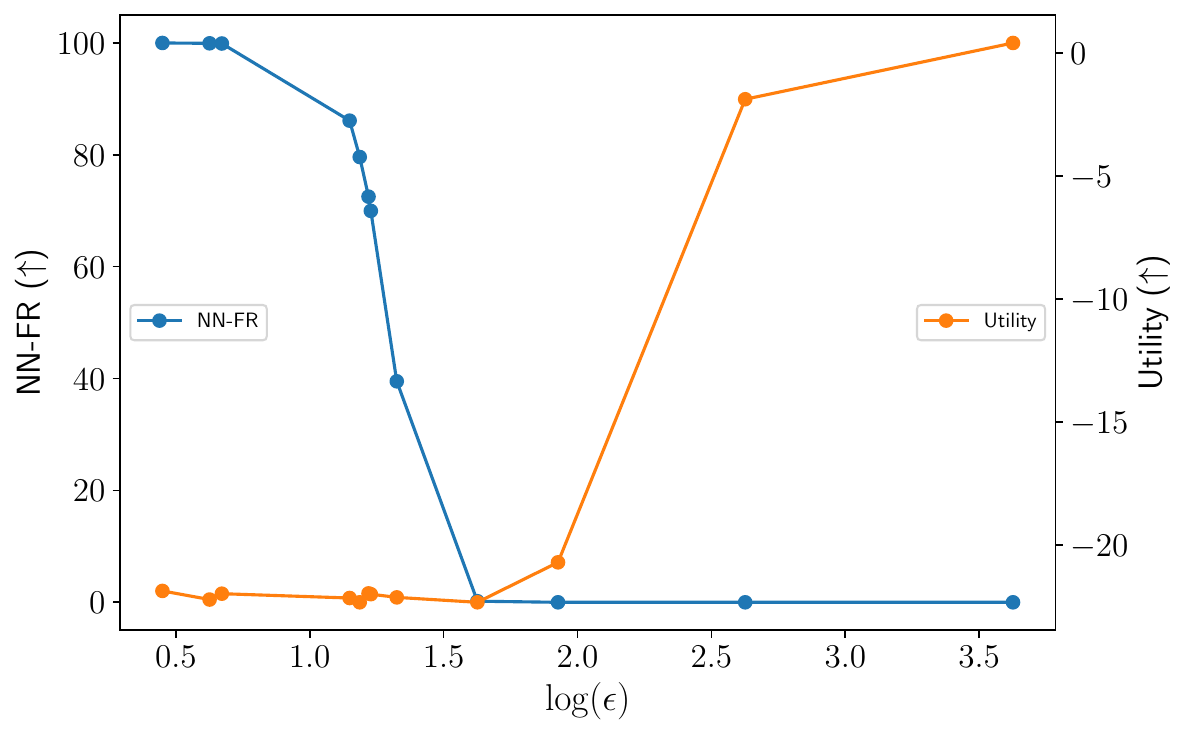}
    \caption{Constant noise (\constnoise) Utility (left, blue) and nearest-neighbor-reconstruction failure rate (right, orange) for increasing standard deviation. Constant noise degrades utility by 20 percentage points compared to the baseline model before beginning to provide any obfuscation.}
    \label{fig:const-noise}
\end{figure}

%--------------------------------------
%   TABLES
%--------------------------------------
\begin{table*}[!t]
% \renewcommand{\arraystretch}{1.3}
% if using array.sty, it might be a good idea to tweak the value of
% \extrarowheight as needed to properly center the text within the cells
    \caption{
    SGT Loss Ablation. Measuring the reducing in performance (Utility), resilience to obfuscations (NN-FR, MRP-FR, SymTTR-10, and SymTTR-100), and information theoretic protection (Hist. Entropy, MI, and PAC-Adv). The \sgtfull \ loss provides strong performance across all three categories.}
    \label{tab:loss-ablation}
% \centering
% Some packages, such as MDW tools, offer better commands for making tables
% than the plain LaTeX2e tabular which is used here.
    \begin{tabular}{l|c|llll|lcl}
    \toprule
    SGT Loss & Utility ($\uparrow$) & NN-FR ($\uparrow$) & MRP-FR ($\uparrow$) & SymTTR-10 ($\uparrow$) & SymTTR-100 ($\uparrow$) & Hist. Entr. ($\uparrow$) & MI ($\downarrow$) & PAC-Adv ($\downarrow$) \\
    \midrule
    \sgtlog & -13.84 & 88.41 & 88.41 & 75.99 & 49.75 & 58.60 & 22.19 & 100.0 \\
    \sgtcos & -5.49 & \textbf{99.98} & 3.21 & 1.43 & 1.07 & 2.47 & 3.2 $\times 10^{19}$ & 100.0 \\
    \sgtabscos & \textbf{-0.01} & 94.90 & \textbf{94.90} & \textbf{93.36} & \textbf{92.86} & 87.62 & 6 $\times 10^9$ & 100.0 \\
    \sgtmi & -0.72 & 34.43 & 34.43 & 18.96 & 17.39 & 23.96 & 1.48 & 16.39 \\
    \sgtmiabscos & -2.44 & 81.23 & 81.23 & 71.36 & 66.88 & 75.72 & 1.91 & 20.56 \\
    \sgtmiabscos  &  &  &  &  &  &  & &  \\
    \ \ \ \ \ + \sgtnormpen& -1.97 & 93.00 & 93.00 & 90.42 & 88.15 & \textbf{91.90} & \textbf{1.11} & \textbf{12.69} \\
    \end{tabular}
\end{table*}

\begin{table*}[!t]
% \renewcommand{\arraystretch}{1.3}
% if using array.sty, it might be a good idea to tweak the value of
% \extrarowheight as needed to properly center the text within the cells
    \caption{
       Performance metrics for large models trained for two epochs on OpenOrca. The SGT manages to maintain high utility while providing strong protection.  }
    \label{tab:big models}
% \centering
% Some packages, such as MDW tools, offer better commands for making tables
% than the plain LaTeX2e tabular which is used here.
    \begin{tabular}{l|r|rrrr|rrr}
    \toprule
    Target Model & Utility ($\uparrow$) & NN-FR ($\uparrow$) & MRP-FR ($\uparrow$) & TTR-50 ($\uparrow$) & SymTTR-100 ($\uparrow$) & Hist. Entropy ($\uparrow$) & MI ($\downarrow$) & PAC-Adv ($\downarrow$) \\
    \midrule
    % Llama-3.3-70B - epoch-0-step-8227 & -0.56 & 98.22 & 98.22 & 98.11 & 98.11 & 80.14 & 0.26 & 3.54 \\
    Llama-3.3-70B & -0.38 & 98.22 & 98.22 & 98.11 & 98.11 & 80.14 & 0.26 & 3.54 \\
    % DeepSeek-R1-Distill-Llama-70B - epoch-1-step-32755 & -0.65 & 97.39 & 97.39 & 97.23 & 97.23 & 78.05 & 0.28 & 3.74 \\
    DeepSeek-R1-70B & -0.46 & 97.39 & 97.39 & 97.23 & 97.23 & 78.05 & 0.28 & 3.74 \\
    % Qwen3-32B-Instruct - epoch-0-step-16456 & -0.30 & 94.95 & 94.95 & 94.54 & 66.39 & 67.67 & 0.23 & 3.03 \\
    Qwen3-32B & -0.29 & 94.95 & 94.95 & 94.54 & 66.39 & 67.67 & 0.23 & 3.03 \\
    % \bottomrule
    \end{tabular}
\end{table*}

\begin{table}[!t]
    \label{tab:pii}
    \caption{PUPA Dataset results. The failure rate of NN and BeamClean reconstructions as well as the PII recovery ratio (PII-Ratio) are reported.}
    \centering
    \begin{tabular}{l|cc|cc}
        \toprule
        SGT Loss& \multicolumn{2}{c|}{FR ($\uparrow$)}  & \multicolumn{2}{c}{PII-Ratio ($\downarrow$)}  \\
         % \midrule
         & NN & BeamClean & NN & BeamClean \\
        \midrule
        \sgtmi & 23.38 & 22.17 & 97.57 & 97.66 \\
        \sgtabscos & 97.67 & 97.64 & 1.60 & 1.61 \\
        \sgtmiabscos & 75.17 & 75.16 & 65.04 & 65.03 \\
        \sgtfull & 86.77 & 86.66 & 46.98 & 47.04 \\
        \bottomrule
    \end{tabular}
\end{table}

% RUN_NAME_MAPPER = {
%     "bright-silence-63": "\sgtfull",
%     "eternal-serenity-56": "\sgtmiabscos",
%     "twilight-rain-57": "\sgtmi",
%     "zesty-oath-29": "\sgtabscos",
%     "youthful-violet-22": "\sgtcos",
%     "fearless-serenity-24": "\sgtlog",
% }

\textbf{Constant noise obfuscations are unable to usefully obfuscate.} Figure \ref{fig:const-noise} plots the Utility and nearest-neighbor reconstruction failure rate (NN-FR) for constant noise obfuscations. This unlearned noise catastrophically degrades model Utility, reducing performance by 20 percentage points compared to baseline for all $\epsilon$ values smaller than 42. At which point utility begins to recover but no privacy is provided as the NN-FR is effectively 0\%.

\textbf{SGTs preserve target model performance.} Table \ref{tab:loss-ablation} shows that in general the SGT does not degrade model performance. The \sgtabscos maintained near identical performance with the baseline, differing only by $0.01$ percentage points (pp). This was followed closely by the \sgtmi \ (-0.72 pp) and \sgtfull \ (-1.97 pp). The exception was \sgtlog which had an over 13 pp drop in performance.
% across all losses the performance on LLM benchmarks compared to the baseline is maintained, with a maximum drop of 5.13\% in LLM utility observed in the \sgtlog utility measurement. The smallest drop in utility is observed in \sgtabscos which was within 0.27 percentage points of the baseline model performance. The full loss, \sgtfull, was within 1.73 percentage points of baseline, beating \sgtmiabscos, showing the value of the median norm penalty.
% \sgtlogcat and \sgtmi, which had the worst performance in terms of attack protection, demonstrating some of the tradeoff between utility and obfuscation performance.
%Utility increases are measured for \sgtlog, \sgtcos, and \sgtmi. This is likely caused by the regularizing effect of the stochastic nature of the SGT during training.

% \textbf{Geometric losses strongly protect against nearest neighbor reconstructions.} The losses that focused on the geometry of obfuscations (\sgtcos and \sgtabscos) provided strong resistance to nearest neighbor attacks, achieving a NN-FR of 97.27\% and 94.90\%, respectively. The full obfuscation loss (\sgtfull) 
\textbf{Nearest Neighbor Reconstructions are an insufficient measure of obfuscation protection.} The top performer in terms of NN-FR was \sgtcos, achieving an impressive 99.98\%. However, the MRP-FR was only 3.21\%. As alluded to in Section \ref{sec:metrics}, this indicates the model is successfully moving the input away from its original location but is laundering information about the embeddings in other ways. This is validated by the histogram entropy (Hist. Entr.). The entropy of the rank histogram of was only 2.47\% of a fully entropic histogram (i.e. a uniform distribution). This together with the less than 2\% SymTTR values implies that the model learned to consistently move directly away from the input, resulting in tokens that were "different" but not truly secure. These results underscore the importance of considering information theoretic measures of privacy beyond just nearest neighbor based reconstructions.

\textbf{Controlling mutual information requires it to be used in the loss.} The losses which did not include the mutual information (i.e., \sgtcos, \sgtcoslog, \sgtlogcat, \sgtabscos)  provided no control over the mutual information. Table \ref{tab:loss-ablation} shows that the geometric based losses involving cosine-similarity had dramatically large mutual information values with \sgtabscos having MI of over 32 quintillion. Additionally, encouraging large standard deviations of components (\sgtlog) provided some control over mutual information but not enough to provide PAC Privacy guarantees. Only the mutual information losses \sgtmi \ , \sgtmiabscos , and \sgtfull \ were able to effectively control the mutual information with the latter providing the best protection against attacks. 

\textbf{The full obfuscation loss provide best protection against attacks.} The \sgtlogcat\ showed weakest protection overall which demonstrates how the approximation of Bound \eqref{eq:gmm-entropy-ub} is not sufficient for privacy protection. The pure \sgtmi~
also performed poorly, though this may have been a result of under-convergence due to hyperparameter sensitivity or being a more difficult optimization problem. The absolute cosine loss (\sgtabscos) provided the best protection in terms of Utility and reconstructions which we tested against. However, as shown with the discrepancy between protection agaisnt NN-FR versus MRP-FR seen in the \sgtcos loss, protection against a specific reconstruction does not imply protection against all. Having high mutual information is a signal that there is information an adversary may be able to exploit. The obfuscation losses that controlled mutual information as these are the only that achieved bounds on feature level reconstruction probability (PAC-Adv), with the full loss achieving 12.69\% bound on reconstruction. 

\textbf{The SGT can protect PII at the text level.} Table \ref{tab:pii} shows the evaluation of the SGT models on the PUPA dataset. All SGTs were able to resist BeamClean with roughly the same strength as against NN. 
Aside from the\sgtmi, all losses were able to maintain strong privacy on the dataset and reduce the recovered PII substantially. The \sgtfull \ decreased recoverable PII by over 50\% (46.98\% vs PII-Ratio NN and 47.04\% vs BeamClean). \sgtabscos \ acheived an impressive reduction of PII recovered by over 90\% (roughly 1.6 PII-Ratio against both reconstructions). The results show that even against more sophisticated reconstructions, the SGT is able to protect private data at the text level and is not simply changing unimportant tokens.

% RUN_NAME_MAPPER = {
%     "bright-silence-63": "\sgtfull",
%     "eternal-serenity-56": "\sgtmiabscos",
%     "twilight-rain-57": "\sgtmi",
%     "zesty-oath-29": "\sgtabscos",
%     "youthful-violet-22": "\sgtcos",
%     "fearless-serenity-24": "\sgtlog",
% }
\textbf{The SGT performs well for large models. }
Table \ref{tab:big models} shows the performance of The SGT for large models. Across all three models the SGT maintains utility to within 0.5pp of their respective baseline models. The SGT for Qwen3 had the highest utility (-0.29pp); however, its SymTTR-100 was lowest (66.39\%) as was its rank histogram entropy (67.67\%). This suggests that further hyperparameter tuning (e.g. increasing the strength of the \sgtmi \ loss) could further improve performance. The remaining metrics are all higher than what was observed in the Llama-1B loss ablation likely due to the substantially longer training time allocated to these models. Overall, the protection offered by the SGT for the large models is high in both transformation metrics and information theoretic measures.

\textbf{MI losses need more time to converge.}
We emphasize that, due to computational constraints, the loss ablation in Table \ref{tab:loss-ablation} was done on a small subset of OpenOrca (15\%) and so some of the performance advantages of \sgtabscos\ come from it being a much simpler optimization problem than $\LossMI$.  As seen in the larger model experiments, Table \ref{tab:big models}, where the \sgtfull \ was trained for a full 2 epochs, the \sgtfull \ loss is able to achieve impressive protection even on transformation metrics. That is, when allowed to converge \sgtfull \ is able to obtain strong transformation metrics with the additional benefit of having theoretical guarantees on reconstruction.
% -------------------------------------
%   Conclustion
% -------------------------------------

\section{Conclusion}\label{sec:conclusion}

In this paper we develop an obfuscation mechanism, the Stained Glass Transform, that information theoretically protects the input embeddings of an LLM at inference time with as little as a 0.29 percentage point drop compared to the LLM's benchmark performance on untransformed embeddings. 
% presented the Stained Glass Transform, an obfuscation mechanism that protects the privacy of inputs to LLMs while preserving the inputs' ability to be used by the model.
We provided a theoretical characterization of the distribution of the SGT across a dataset as well as details of how to perform inference and train an SGT model for a target LLM. Using the characterization of the SGT's distribution across the dataset we derived an obfuscation loss that reduces the mutual information (MI) between input embeddings and their obfuscations. We then provided experimental evidence that, while losses based on common MI bounds provide some control over MI, having the MI in the loss was the only way to obtain information theoretic guarantees of protection.
% showed losses which did not have this mutual information component generally provided weaker protection against our proposed attacks and these losses uniformly failed to provide theoretical guarantees of protection. 
Finally, we showed that the SGT generalizes across model architectures and sizes, maintaining their target model's utility while providing strong protection.


% \newpage
% \IEEEtriggeratref{8}


\bibliographystyle{IEEEtran}
\bibliography{references}
% \bibliography{IEEEabrv,../bib/paper}


% \end{thebibliography}

\end{document}